\documentclass{article}
\usepackage{spconf,amsmath,graphicx}
\usepackage[tight,footnotesize]{subfigure}
\usepackage{multirow}
\usepackage{rotating}
\usepackage{xcolor,colortbl}
\usepackage{amsmath}
\usepackage{amsfonts}
\usepackage{amssymb}
\usepackage{color}
\usepackage{epsfig}
\usepackage{psfrag}
\usepackage{epstopdf}
\usepackage{booktabs}

\title{Frequency Selective Extrapolation with Residual Filtering for Image Error Concealment}

\name{J\'an Koloda$^{\dagger}$, J\"{u}rgen Seiler$^{\ddagger}$, Andr\'{e} Kaup$^{\ddagger}$,  Victoria S\'anchez$^{\dagger}$ and Antonio M. Peinado$^{\dagger}$\thanks{This work has been supported by an FPU grant from the Spanish Ministry of Education and by the MICINN TEC2010-18009 project.}}

\address{
\begin{minipage}{7.5cm}
\centering
\small  $^\dagger$Dpt.\ of Signal Theory, Networking and Communications \\[1mm]
\small  Universidad de Granada, Spain \\
\small \{janko, victoria, amp\}@ugr.es
\end{minipage}
\begin{minipage}{8.5cm}
\centering
\small $^{\ddagger}$Chair of  \ Multimedia Communications and Signal Processing \\[1mm]
\small University of Erlangen-Nuremberg, Germany \\
\small \{seiler, kaup\}@LNT.de
\end{minipage}
}

\begin{document}
%

\maketitle
\begin{abstract}
The purpose of signal extrapolation is to estimate unknown signal parts from known samples. This task is especially important for error concealment in image and video communication.
For obtaining a high quality reconstruction, assumptions have to be made about the underlying signal in order to solve this underdetermined problem.
Among existent reconstruction algorithms, frequency selective extrapolation (FSE) achieves high performance by assuming that image signals
can be sparsely represented in the frequency domain.
However, FSE does not take into account the low-pass behaviour of natural images. In this paper, we propose a modified FSE that takes
this prior knowledge into account for the modelling, yielding significant PSNR gains. 
\end{abstract}

\section{Introduction}
\label{sec:intro}

Signal reconstruction is a very challenging task for many multimedia applications where the quality of the received data is of utmost importance. A common example
is the transmission of image/video signals over error prone channels which may yield block losses. The lost areas need to be concealed employing the information
provided by the correctly received data. There are several examples of efficient error concealment (EC) techniques applied to image communication.
The EC algorithm proposed in \cite{Markov} is based on Markov random fields and
focuses on preserving visually important features, such as edges.
Bilateral filtering that exploits a pair of gaussian kernels is treated in \cite{BLF}. In \cite{SLP}, the lost region is recovered through sparse linear prediction.
Moreover, inpainting \cite{inpaintTH} can also be employed for concealment purposes.

An alternative approach to image EC is the frequency selective extrapolation (FSE) proposed in \cite{FSE_AEUE}.
In particular, the complex-valued FSE implementation \cite{FSE_SP_letters}
can provide high quality reconstructions with a low computational burden. This technique develops a signal model from the set of Fourier
basis functions which can be used to replace the unknown samples. Although this FSE algorithm basically consists in determining frequency components,
it does not exploit any a priori knowledge regarding the typical spectrum of natural images, which may result in high-frequency artifacts.
In this paper, we propose the introduction of a low-pass filtering in the FSE iterative procedure which can efficiently account for this fact,
increasing the FSE performance while maintaining a low computational cost.

The paper is organized as follows. In Section \ref{sec:fse_overview}, we provide a short review of the FSE algorithm. Our proposal, based on residual filtering, is described
in Section \ref{sec:mod_fse}. Experimental results are discussed in Section \ref{sec:results}. The last section is devoted to conclusions.

\section{Frequency Selective Extrapolation}
\label{sec:fse_overview}

Our proposal is a modification of the complex-valued implementation of FSE  \cite{FSE_SP_letters}.
This approach is able to robustly reconstruct various image contents at very high quality \cite{FSE_SP_letters, FSE_AEUE, FFSE_ortho}.
We briefly summarize it in this section.

\begin{figure}
\centering
\includegraphics[width=3.20in]{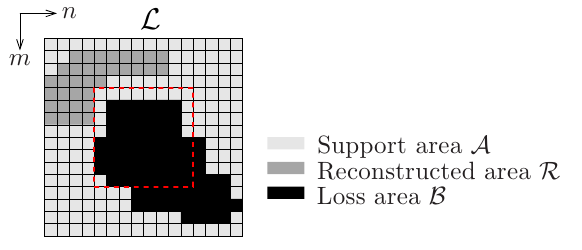}
\caption[Areas]{\small{Extrapolation area $\mathcal{L}$ as union of support area $\mathcal{A}$, reconstructed area $\mathcal{R}$, and loss area $\mathcal{B}$.
The currently processed block (marked by the red dashed line) is located in the centre.}}
\label{fig:extrapolation_area}
\end{figure}

During the extrapolation process of FSE, the image is divided into blocks of equal size. Besides the block actually containing areas to be reconstructed,
neighbouring samples belonging to adjacent blocks are taken into account, as well. All the considered samples make up the so called extrapolation area $\mathcal{L}$
(an example is shown in Fig.\ \ref{fig:extrapolation_area}). The size of area $\mathcal{L}$ is $M\times N$ samples and the signals in this area are indexed by spatial
variables $m$ and $n$. All samples in area $\mathcal{L}$ belong to one of the three following groups: the known samples built up support area $\mathcal{A}$,
all unknown samples belong to the loss area $\mathcal{B}$ (located at the centre of $\mathcal{L}$)
and all samples from neighbouring blocks that have been extrapolated before belong to the reconstructed area $\mathcal{R}$.

FSE extrapolation is carried out from a parametric model
\begin{equation}
g(m,n) = \sum_{\left(k,l\right)\in\mathcal{K}}c_{k,l} \varphi_{k,l}(m,n) .
\end{equation}
This is a weighted superposition of two-dimensional basis functions $\varphi_{k,l}\left(m,n\right)$ with weights $c_{k,l}$. In this work we will employ Fourier functions,
\begin{equation}
\varphi_{k,l}(m,n) = \frac{1}{MN}\mathrm{e}^{\frac{2\pi\mathrm{j}}{M}km} \mathrm{e}^{\frac{2\pi\mathrm{j}}{N}ln}.
\end{equation}
As described in detail in \cite{FSE_SP_letters}, the model generation is performed iteratively, with the initial model $g^{\left(0\right)} (m,n)$ being $0$, which involves
that coefficients $c_{k,l}^{(0)}$ are also set to $0$.
At every iteration, one of the possible basis functions is selected. After estimating the corresponding weight, it is added to the model that has been generated so far.
In order to determine the best basis function and its weight at every iteration $\nu$, the residual
\begin{equation}
r^{\left(\nu\right)} (m,n) = \left( s(m,n) - g^{\left(\nu\right)} (m,n) \right)
\cdot b(m,n)
\end{equation}
between the available signal $s(m,n)$ and the current model $g^{\left(\nu\right)} (m,n)$ generated so far is regarded. Window $b(m,n)$ is zero for $(m,n)\in\mathcal{B}$
and one otherwise in order to ensure that unknown pixels are not used.

The best function $\varphi_{u,v}(m,n)$ at this iteration, conveniently weighted by a factor $\Delta c_{u,v}$, will be the one which can better approximate this residual.
Let us suppose that we already know this function. Then, the corresponding model coefficient will be updated as
\begin{equation}
c_{u,v}^{(\nu+1)}=c_{u,v}^{(\nu)} + \gamma\Delta c_{u,v}
\label{dc_gamma}
\end{equation}
and the residual for the next iteration will be
\begin{equation}
r_{u,v}^{(\nu+1)} (m,n) = \left( r^{(\nu)} (m,n) -\Delta c_{u,v} \varphi_{u,v}(m,n) \right)
\cdot b(m,n).
\label{res_next}
\end{equation}
Factor $\gamma$ in Eq.(\ref{dc_gamma}) is introduced to compensate the orthogonality deficiency of the proposed framework \cite{FFSE_ortho}.
Coefficient $\Delta c_{u,v}$ is estimated by minimizing a weighted square error obtained from this last residual as
\begin{equation}
E_{u,v}^{(\nu+1)} = \sum_{(m,n)\in\mathcal{L}} w(m,n) \left| r_{u,v}^{(\nu+1)} (m,n) \right|^2.
\end{equation}
Finally, the desired coefficient is
\begin{equation}
\Delta c_{u,v} = \frac{\displaystyle \sum_{(m,n)\in\mathcal{L}} r^{(\nu)} (m,n) \varphi^\ast_{u,v} (m,n) w(m,n)}{\displaystyle \sum_{(m,n)\in\mathcal{L}} \varphi_{u,v}^\ast(m,n)w(m,n)\varphi_{u,v}(m,n)}
\label{proj_coeff} 
\end{equation}
which can be interpreted as a weighted projection coefficient of $r^{(\nu)} (m,n)$ on $\varphi_{u,v}(m,n)$,
The weighting function $w(m,n)$ can be defined as \cite{FSE_AEUE}
\begin{equation}
w(m,n) = \left\{\begin{array}{ll} \hat{\rho}^{\sqrt{\left(m-\frac{M-1}{2}\right)^2+\left(n-\frac{N-1}{2}\right)^2}} & \forall \left(m,n\right)\in \mathcal{A} \\ \delta\hat{\rho}^{\sqrt{\left(m-\frac{M-1}{2}\right)^2+\left(n-\frac{N-1}{2}\right)^2}} & \forall \left(m,n\right)\in \mathcal{R} \\ 0 & \forall \left(m,n\right)\in \mathcal{B} \end{array} \right. .
\end{equation}
Using this function, the influence of each sample on the model generation can be controlled according to its position.
This is also the reason why the weighting function is divided into three different parts. As all unknown samples cannot contribute to the model generation,
they have to be excluded from the calculations. Accordingly, their weight in area $\mathcal{B}$ is set to $0$. For the known samples, an exponentially decaying weight is
used for reducing their influence with increasing distance to the area to be extrapolated in the current block. Parameter $\hat{\rho}$ controls the speed of the decay.
As samples from neighbouring blocks that are originally not known but have been extrapolated before are not as reliable as originally available samples, the influence
for these samples is weighted by an extra factor $\delta \in [0, 1]$.

The remaining issue is the determination of the best function $\varphi_{u,v}(m,n)$.
In order to do so, we must consider that, in fact, the projection coefficient and the square error can be computed for every basis function $\varphi_{k,l}(m,n)$.
Furthermore, considering the orthogonality principle, every square error $E_{k,l}^{(\nu+1)}$ can be decomposed as the square error determined for the previous iteration $E^{(\nu)}$
minus the achieved decrease of square error, which is defined as \cite{FSE_AEUE},
\begin{equation}
\Delta E_{k,l}^{(\nu)}  = \left|\Delta c_{k,l} \right|^2 \mkern-10mu \sum_{(m,n)\in\mathcal{L}}
\mkern-10mu \varphi_{k,l}^\ast (m,n) w(m,n)\varphi_{k,l} (m,n).
\end{equation}
The basis function can be selected now as the one which maximizes this decrease, that is,
\begin{equation}
 (u,v) = \mathop{\mathrm{argmax}}_{(k,l)} \Delta E_{k,l}^{(\nu) }.
\end{equation}

After the model generation has finished, all the samples that are originally not known are taken from the model and inserted at the corresponding positions of the incomplete original signal. 

\section{FSE with Residual Filtering}
\label{sec:mod_fse}
It is well known that low frequencies are likely to yield larger Fourier coefficients than high ones in natural images \cite{Isotropic_Images, Recom}.
This is an a priori knowledge not considered in the original
FSE algorithm which could be incorporated into it in order to improve both reconstruction quality and robustness.
Thus, in the same way as the knowledge about spatial influence is controlled with weights $w(m,n)$, we propose here the use of a frequency weighting (filtering) which, applied to the
residuals, can exploit this a priori knowledge about frequency importance.
In order to do so, it is convenient to express both residuals and square errors in the frequency domain.

\subsection{FSE in the frequency domain}
FSE can be efficiently implemented and easily viewed in the frequency domain \cite{FSE_SP_letters}.
Let us consider the spatially-weighted version of the residual,
\begin{equation}
r_w^{(\nu)}(m,n) = w(m,n) r^{(\nu)}(m,n).
\end{equation}
Then, from Eq. (\ref{proj_coeff}), the projection coefficient for function $\varphi_{k,l}(m,n)$ can be expressed as
\begin{equation}
\Delta c_{k,l} = MN\frac{R_w^{(\nu)}(k,l)}{W(0,0)},
\label{proj_coeff_freq}
\end{equation}
where $R_w^{(\nu)}(k,l)$ and $W(k,l)$ are the DFTs of $r_w^{(\nu)}(m,n)$ and $w(m,n)$, respectively.
Also, the decrease of square error can be expressed as,
\begin{equation}
\Delta E_{k,l}^{(\nu)} = \frac{|R_w^{(\nu)}(k,l)|^2}{W(0,0)}.
\label{decrease_freq}
\end{equation}
Finally, from Eq. (\ref{res_next}), it is easily deduced that
\begin{equation}
R_w^{(\nu+1)}(k,l) = R_w^{(\nu)}(k,l) - \frac{1}{MN}\Delta c_{u,v}W(k-u,l-v),
\label{update_freq}
\end{equation}
which provides the weighted residual required for the next iteration directly in the frequency domain.
Equations (\ref{proj_coeff_freq})-(\ref{update_freq}) provide an efficient implementation of FSE, since it can be fully carried out in the frequency domain.

\begin{figure}
\centering
\includegraphics[width=3.00in]{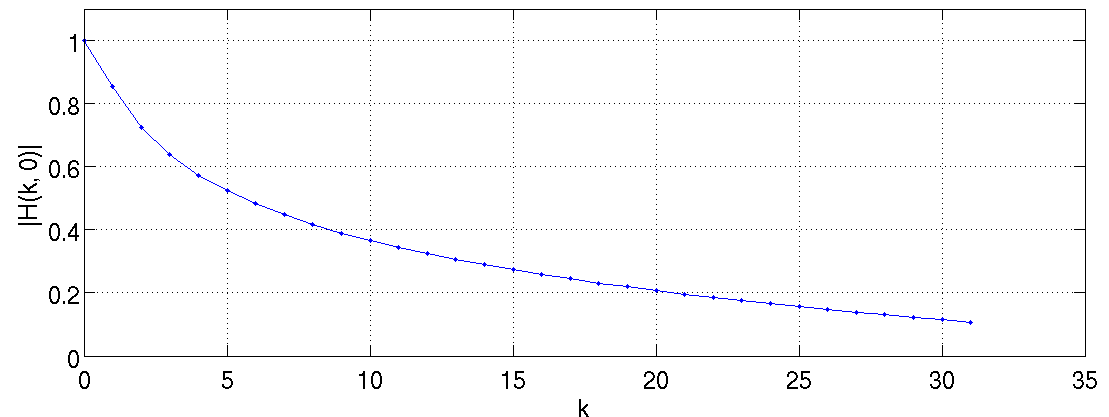}
\caption[Filter]{\small{One-dimensional profile of the filter $H$ of size 64$\times$64 with $f_0 = 0.0098$ and $G = 292.9$.}}
\label{fig:rps}
\end{figure}

\subsection{Filtering the weighted residual (XFSE)}
We can see that the evolution of the iterative procedure relies on the computation carried out in Eqs. (\ref{proj_coeff_freq}) and (\ref{decrease_freq}),
that is, on the weighted residual $R_w^{(\nu)}(k,l)$.
Therefore, a possible way of incorporating the a priori knowledge about the low-pass behaviour of natural images can be the low-pass filtering of the residual in these equations, that is,
\begin{equation}
\Delta c_{k,l} = MN\frac{R_w^{(\nu)}(k,l)H(k,l)}{W(0,0)},
\end{equation}
\begin{equation}
\Delta E_{k,l}^{(\nu)} = \frac{|R_w^{(\nu)}(k,l)H(k,l)|^2}{W(0,0)},
\end{equation}
where $H(k,l)$ is even, real-valued and non-negative, and represents the frequency response of the applied low-pass filter. The rest of the procedure can be kept unaltered.
The resulting procedure will be referred to as XFSE in the following.

The main issue to be addressed now is the low-pass filter selection. After some preliminary experiments, we have applied a filter with the following circularly symmetric frequency response,
\begin{equation}
H(k,l)=\frac{\log\left[ G\frac{f_0}{2\pi} \frac{1}{\left[ f_0^2 + \left( \frac{k}{M}\right)^2+\left( \frac{l}{N}\right)^2 \right]^{3/2}} \right]}
{\log\left( \frac{G}{2\pi f_0^2} \right)}.
\end{equation}
This filter is inspired on the average power spectral density of natural (isotropic) images given in \cite{Isotropic_Images}, modified with a gain factor $G$, smoothed by logarithm,
and normalized to provide $H(0,0)=1$. Parameter $f_0$ controls the bandwidth.
A one-dimensional profile of this filter is shown in Fig. \ref{fig:rps}.

\begin{figure}
\centering
\subfigure[]{
\includegraphics[width=1.60in]{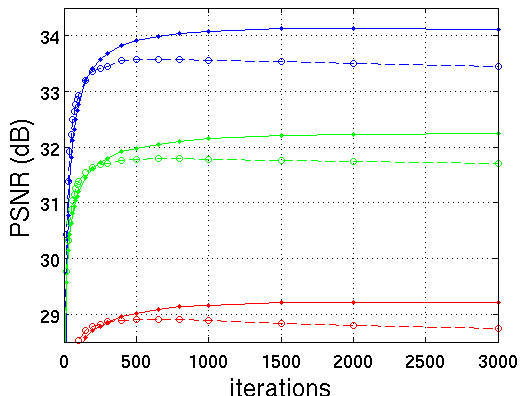}
\label{fig:psnr_errors}
}
\subfigure[]{
\includegraphics[width=1.60in]{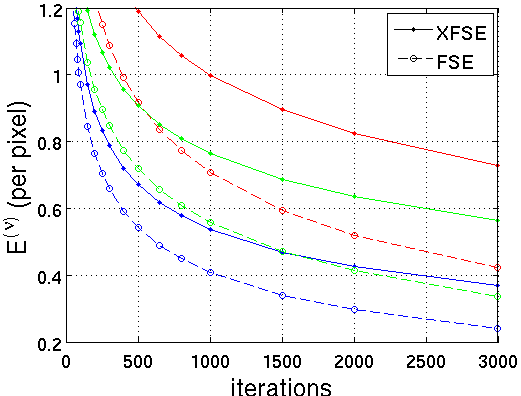}
\label{fig:error}
}
\caption[dE]{\small{Performance overview in terms of \subref{fig:psnr_errors} PSNR and \subref{fig:error} residual energy $E^{\left(\nu\right)}$ of FSE and XFSE
for the images of \textit{Peppers} (blue), \textit{Boat} (red) and \textit{Goldhill} (green). Dispersed error pattern is employed.}}
\label{fig:dE}
\end{figure}

Let us analyze now the effect of this filter over the square error decrease.
At every FSE iteration, the basis function that produces the largest decrease in the residual energy $\Delta E_{k,l}^{(\nu)}$ is selected.
However, this may lead to overfitting since the reconstruction quality decreases once a critical number of iterations is achieved \cite{FFSE_ortho}, while the weighted residual
error $E^{(\nu)}$ keeps falling (see Fig.\ref{fig:psnr_errors}).
In order to prevent this overfitting, when several basis functions yield a comparable (maximum) decrease $\Delta E_{k,l}^{(\nu)}$, the introduced filtering favours the lowest frequencies.
This is illustrated in Fig. \ref{fig:error}, where we can see that XFSE yields higher weighted residual error $E^{(\nu)}$ but, however, improves the reconstruction quality.

Regarding the projection coefficient $\Delta c_{u,v}$ for the selected function, since $H(k,l)\le 1$, the filter acts as a weighting factor that reduces the contribution of high
frequencies to the reconstructed signal. This does not mean that high frequencies are avoided, since if a high frequency is a clear candidate to be included in the signal model,
this frequency will appear again in subsequent iterations. However, if it is not, it will only appear spuriously, and its contribution to the final signal model will be negligible.

Since $H(u,v) \leq 1$, we can alternatively see our filtering as a dynamic reduction of the orthogonality deficiency compensation factor $\gamma$.
As shown in \cite{FFSE_ortho}, smaller compensation factors yield a better convergence
(slower performance decrease after a certain number of iterations) although
more iterations are required to achieve maximum performance.
Although we will frequently find that $H(u,v)\gamma \ll \gamma$, during the first iterations low frequencies
with high $H(k,l)$ tend to be selected, so there will be only little penalization in reconstruction quality. On the other hand, in later iterations higher frequencies
are selected in order to tune fine details.
For these frequencies, the effective orthogonality deficiency compensation factor $H(u,v)\gamma$ is smaller and the convergence is improved as remarked above.
This is shown in the next section.

\begin{figure}
\centering
\subfigure[]{
\includegraphics[width=0.75in]{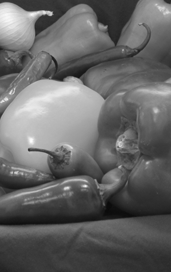}
\label{fig:peppers_orig}
}
\subfigure[]{
\includegraphics[width=0.75in]{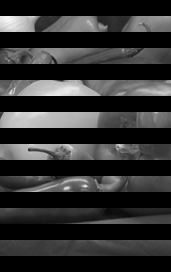}
\label{fig:peppers_mask}
}
\subfigure[]{
\includegraphics[width=0.75in]{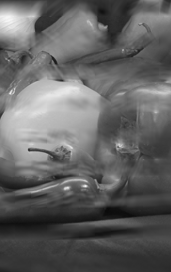}
\label{fig:peppers_fse}
}
\subfigure[]{
\includegraphics[width=0.75in]{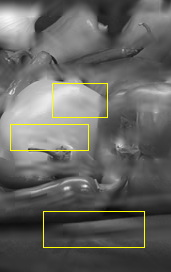}
\label{fig:peppers_rhoS}
}
\caption[Results]{\small{Subjective comparison for a fraction of \textit{Peppers}. \subref{fig:peppers_orig} Original image. \subref{fig:peppers_mask} Received data.
\subref{fig:peppers_fse} Reconstruction by FSE. \subref{fig:peppers_rhoS} Reconstruction by XFSE.}}
\label{fig:results}
\end{figure}

\section{Experimental results}
\label{sec:results}

The performance of our proposal is tested on the images of $\textit{Peppers}$ (384$\times$512), $\textit{Boat}$ (512$\times$512) and $\textit{Goldhill}$ (720$\times$576).
In addition, the set of 24 images (768$\times$512) by Kodak \cite{Kodak} is also used. We will employ a dispersed error pattern with a block loss rate of around 25\% (see \cite{SLP}
for details). In addition, consecutive block losses (50\% loss rate) will also be considered (see Fig. \ref{fig:peppers_mask}).ç
The blocks are considered to have dimensions of 16$\times$16 pixels and the size of $\mathcal{L}$ is 48$\times$48.
We compare the performance with other spatial EC techniques,
namely EC based on Markov random field (MRF) \cite{Markov}, inpainting (INP) \cite{inpaintTH}, bilateral filtering (BLF) \cite{BLF} and sparse linear prediction (SLP) \cite{SLP}.

To set up the filter, the gain factor $G$ has been heuristically set to 292.9 in order to guarantee that the filter frequency response is always positive.
On the other hand, the filter bandwidth is usually expressed as $f_0 = \alpha/2\pi$ and the value of $\alpha$ is around 0.06 \cite{Isotropic_Images} leading to $f_0 = 0.0098$ which
involves a 3dB-cutoff bin of $2.17$ for $N=M=64$. The remaining FSE parameters are set according to \cite{FSE_SP_letters}, with $\gamma = 0.25$.

A comparison of XFSE and FSE is shown in Fig.\ref{fig:results}. By applying the residual filtering, the performance is improved on average by approximately 0.4dB. This improvement
is even higher when consecutive block losses are considered. Also, it is observed that
the performance decrease with high a number of iterations is alleviated. Note that although XFSE achieves the maximum performance using more iterations than FSE, XFSE already outperforms
FSE at the number of iterations for which FSE reaches its maximum PSNR. This behaviour is also reflected in Table \ref{tab:results}.

\begin{table}
\scriptsize
  \begin{tabular}{m{0.55cm}m{0.1cm}|m{0.5cm}m{0.5cm}m{0.5cm}m{0.5cm}m{0.7cm}m{0.7cm}m{0.7cm}}
\hline    \hline
    &					&		MRF&	INP&	BLF&	SLP&	FSE$_{\textit{max}}$&	XFSE$_{\textit{fse}}$& 	XFSE$_{\textit{max}}$ 	\\ \hline
    \multirow{2}{*}{\textit{Peppers}}&	(a)&		32.59&	33.13&	33.17&	33.94&	33.58&			33.91&		\textbf{34.13} 			\\
    &					(b)&		25.04&	25.28&	25.43&	24.64&	25.47&			26.18&		\textbf{26.24} 			\\ \hline
    \multirow{2}{*}{\textit{Boat}}&	(a)&		27.91&	27.79&	28.37&	28.54&	28.90&			29.02&		\textbf{29.22} 			\\
    &					(b)&		23.07&	22.69&	22.85&	22.48&	23.75&			23.97&		\textbf{24.16} 			\\ \hline
    \multirow{2}{*}{\textit{Goldhill}}&	(a)&		31.12&	30.40&	30.91&	31.72&	31.79&			32.10&		\textbf{32.24} 			\\
    &					(b)&		26.09&	25.82&	24.49&	26.19&	26.56&			27.00&		\textbf{27.05} 			\\ \hline
    \multirow{2}{*}{Kodak}&		(a)&		29.61&	28.76&	29.64&	29.92&	30.45&			30.54&		\textbf{30.69} 			\\
    &					(b)&		24.76&	24.38&	24.83&	24.84&	25.30&			25.63&		\textbf{25.71} 			\\ \hline \hline
  \end{tabular}  
 \caption[PSNR] {\small{PSNR values
(in dB, whole images) for test images reconstructed by several algorithms. The average PSNR for the Kodak set is also included. Dispersed error pattern (a) and consecutive losses (b) are applied.
The best performances are in bold face.}}   
\label{tab:results}
\end{table}

Table \ref{tab:results} shows a PSNR comparison of the tested techniques for dispersed and consecutive losses.
The best performance of FSE (FSE$_{\textit{max}}$) is compared to the best performance of XFSE (XFSE$_{\textit{max}}$)
as well as to XFSE using the same number of iterations as FSE$_{\textit{max}}$ (XFSE$_{\textit{fse}}$).
Our proposal outperforms other state-of-the-art techniques and improves
the reconstruction quality with respect to FSE by up to 0.5dB  for dispersed losses and 0.7dB for consecutive losses.
Finally, simulations reveal that the processing time is increased by approximately only 13\%.

\begin{figure}
\centering
\subfigure[]{
\includegraphics[width=1.60in]{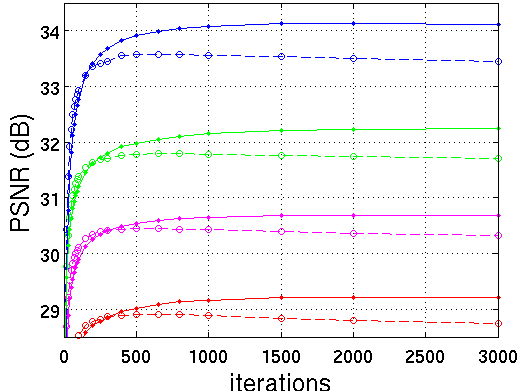}
\label{fig:dispersed}
}
\subfigure[]{
\includegraphics[width=1.60in]{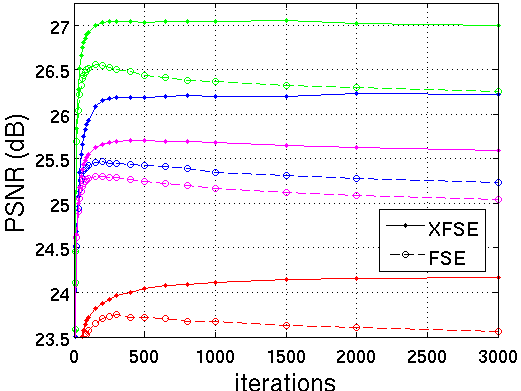}
\label{fig:consecutive}
}
\caption[PSNR]{\small{Performance comparison for \subref{fig:dispersed} dispersed and \subref{fig:consecutive} consecutive losses. The PSNR for
\textit{Peppers} (blue), \textit{Boat} (red), \textit{Goldhill} (green) and the average PSNR for the Kodak set (magenta) are shown.
}}
\label{fig:psnr}
\end{figure}

\section{Conclusions}
\label{sec:conclusions}
We have proposed the introduction of the prior knowledge about the natural image spectra into the FSE algorithm.
This is achieved by filtering the residual error by a specifically designed low-pass filter. Better convergence  and gains of up to 0.7dB with respect to the original FSE
are achieved with a marginal additional computational cost.

\bibliographystyle{IEEEbib}
\bibliography{strings,refs,books}

\end{document}